\documentclass[runningheads]{llncs}
\usepackage[T1]{fontenc}
\usepackage{graphicx}
\usepackage{hyperref}

\usepackage{color}

\newcommand{\ours}[0]{FACT}
\usepackage{tikz}
\usepackage{makecell}
\usepackage{booktabs}
\usepackage{amsmath}
\usepackage{xspace}
\usepackage{subcaption}
\usepackage{amssymb}
\usepackage{lipsum}
\usepackage[export]{adjustbox}
\usepackage{microtype}

\let\titleold\title
\renewcommand{\title}[1]{\titleold{#1}\newcommand{\thetitle}{#1}}
\def\maketitlesupplementary
   {
   \newpage
       \onecolumn
        \textbf{\Large \centering{Supplementary Material}}\\
   }

\AtEndPreamble{
    \usepackage[capitalize]{cleveref}
    \crefname{section}{Sec.}{Secs.}
    \Crefname{section}{Section}{Sections}
    \Crefname{table}{Table}{Tables}
    \crefname{table}{Tab.}{Tabs.}
}

\makeatletter
\DeclareRobustCommand\onedot{\futurelet\@let@token\@onedot}
\def\@onedot{\ifx\@let@token.\else.\null\fi\xspace}

\def\eg{\emph{e.g}\onedot} 
\def\ie{\emph{i.e}\onedot}

\def\etal{\emph{et al}\onedot}
\makeatother

\begin{document}
\title{FACT: Multinomial Misalignment Classification for Point Cloud Registration}

\author{Ludvig Dillén\inst{1}\orcidID{0009-0009-2946-2990} \and
Per-Erik Forssén\inst{2}\orcidID{0000-0002-5698-5983} \and
Johan Edstedt\inst{2}\orcidID{ 0000-0002-1019-8634}}
\authorrunning{L. Dillén et al.}
\institute{Centre for Mathematical Sciences, Lund University, Lund, Sweden
\email{ludvig.dillen@math.lth.se}\\
 \and
Department of Electrical Engineering, Linköping University, Linköping, Sweden\\
\email{\{per-erik.forssen, johan.edstedt\}@liu.se}}

\maketitle              % typeset the header of the contribution
\begin{abstract}
We present \ours, a method for predicting alignment quality (\ie, registration error) of registered lidar point cloud pairs.
This is useful \eg\ for quality assurance of large, automatically registered 3D models.
\ours~extracts local features from a registered pair and processes them with a point transformer-based network to predict a misalignment class. We generalize prior work that study binary alignment classification of registration errors, by recasting it as multinomial misalignment classification.
To achieve this, we introduce a custom regression-by-classification loss function that combines the cross-entropy and Wasserstein losses, and demonstrate that it outperforms both direct regression and prior binary classification.
\ours~successfully classifies point-cloud pairs registered with both the classical ICP and GeoTransformer, while other choices, such as standard point-cloud-quality metrics and registration residuals are shown to be poor choices for predicting misalignment.
On a synthetically perturbed point-cloud task introduced by the CorAl method, we show that \ours~achieves substantially better performance than CorAl.
Finally, we demonstrate how \ours~can assist experts in correcting misaligned point-cloud maps. Our code is available at \url{https://github.com/LudvigDillen/FACT_for_PCMC}.

\end{abstract}

\section{Introduction}\label{sec:introduction}
We propose \textit{Feature-Aware Classification Transformer} (FACT), a method for lidar {\it Point Cloud Misalignment Classification} (PCMC), \ie, for predicting if a pair of point clouds has been correctly aligned. The process of aligning point cloud pairs is extensively studied under the name {\it point cloud registration} (PCR)~\cite{huang2021comprehensive,almqvist2018learning,tavares20,li2021tutorial} and consists of finding a rigid transformation that accurately places a pair in a common coordinate system. Point cloud registration is a basic component of point cloud-based mapping~\cite{moosmann11,zhang_rss14} where lidar point clouds are incrementally added to a 3D model. As lidar scans are successively registered, an increasingly detailed pose trajectory and a map are obtained, for use in applications such as autonomous driving~\cite{geiger12}, underground mining~\cite{magnusson07}, building information modeling~\cite{huang2021comprehensive,chen2013point}, and terrain mapping~\cite{rostain2024}.

\begin{figure}[t!]
    \centering
    \hspace{-1.2cm} % Shift the figure slightly left
    \begin{minipage}{0.5\columnwidth}
        \centering
        \begin{tikzpicture}
        \node[anchor=south west] (main) at (0,0) {\includegraphics[width=0.60\linewidth, trim={2cm 2cm 4cm 2.5cm}, clip]{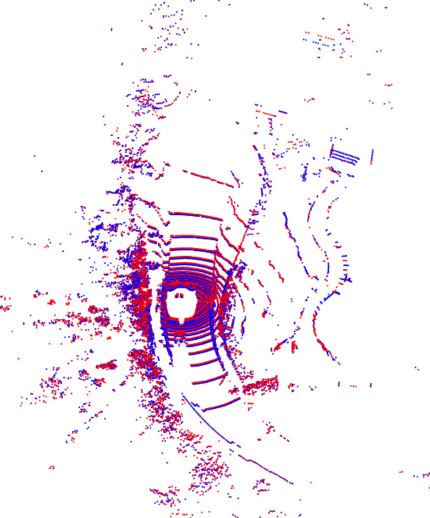}};
        \node[anchor=north] at (0.2,4.4) {\parbox{2.3cm}{\centering \scriptsize Predicted Class 0\\GT Class 0}};
        \end{tikzpicture}
    \end{minipage}%
    \hspace{-0.5cm} % Shift the figure slightly left
       \begin{minipage}{0.5\columnwidth}
        \centering
        \begin{tikzpicture}
        \node[anchor=south west] (main) at (0,0) {\includegraphics[width=0.60\linewidth, , trim={4cm 2cm 2cm 2.5cm}, clip]{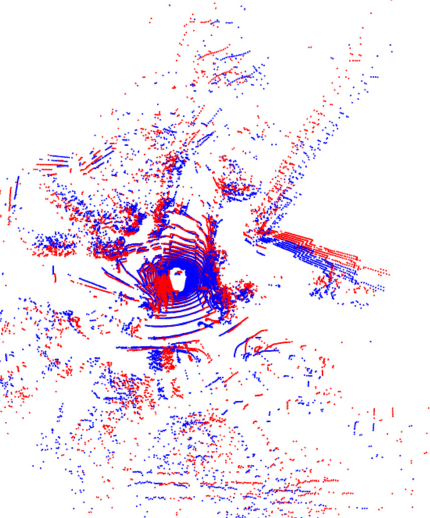}};
        \node[anchor=south west] (zoomed) at (1.4,2.9) {\includegraphics[width=0.45\linewidth]{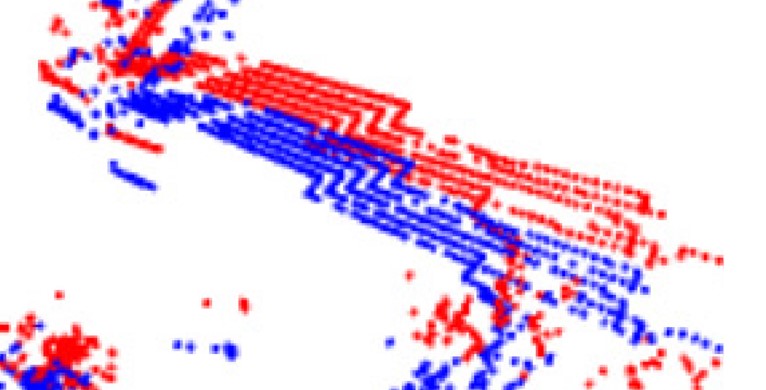}};
        \draw[thick, red] (1.5,3.0) rectangle ++(2.6,1.4);
        \draw[thick] (2.8,2.6) -- (2.8,3.0);
        \draw[thick, red] (2.2,1.9) rectangle ++(1.2,0.7);
        \node[anchor=north] at (0.2,4.4) {\parbox{2.3cm}{\centering \scriptsize Predicted Class 9\\GT Class 9}};
    \end{tikzpicture}
    \end{minipage}
    \caption{
    Two examples of point clouds registrations from nuScenes~\cite{nuscenes}. Left: a correctly aligned pair; right: a misaligned pair (see red box). \ours~easily finds the error class of the synthetically perturbed pairs.
    This paper proposes a method for predicting the quality (registration error) of a given point cloud alignment, both for synthetic errors, as shown here, and real registered point clouds pairs.}
    \label{fig:intro_example}
\end{figure}
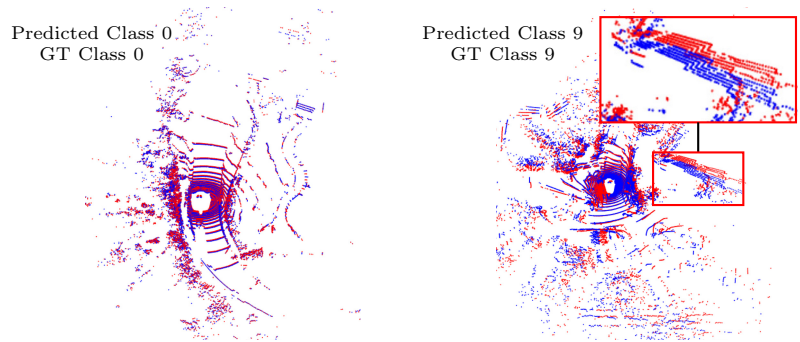

During 3D mapping, registration errors inevitably occur, resulting in drift and map errors. This is problematic as it leads to inaccurate assumptions about the physical world, causing flawed decision-making in downstream tasks.
Research on improved map accuracy is usually directed towards improving the registration method, either by improving the algorithm or by adapting the method to a specific use case.
A less explored approach is to validate that the registration estimates are correct, \ie, perform PCMC. There are several reasons why this task is relevant. 
When a registration algorithm is used in an offline setting, it is commonly followed by manual visual inspection. Our method can replace or facilitate this inspection, by highlighting likely errors.
This concept is particularly valuable in machine learning systems as it can serve as a self-supervision tool. 
PCMC can be used for self-supervision in registration pipelines both for training and automatic correction of registration errors.

In \cref{fig:intro_example}, we give an example of an aligned and a misaligned point cloud pair. Previous works on PCMC ~\cite{almqvist2018learning,adolfsson2021coral,droeschel2018efficient,fujii2015detection,nobili2018predicting,yin2019failure,akai2022detection,bogoslavskyi2017analyzing,camous2022deep}
have been mainly focused on specific in-house SLAM pipelines, whereas we aim to develop a standalone model, applicable to a more general setting. Closest to our work is CorAl~\cite{adolfsson2021coral} which thus serves as our baseline. Different from our method, all previous work is limited to binary classification (\ie\ {\it aligned} or {\it misaligned}).

\begin{figure}[t]
    \centering
    \includegraphics[width=\linewidth, trim={1.35cm 0.10cm 0.22cm 1.13cm}, clip, max size={\linewidth}{\textheight}]{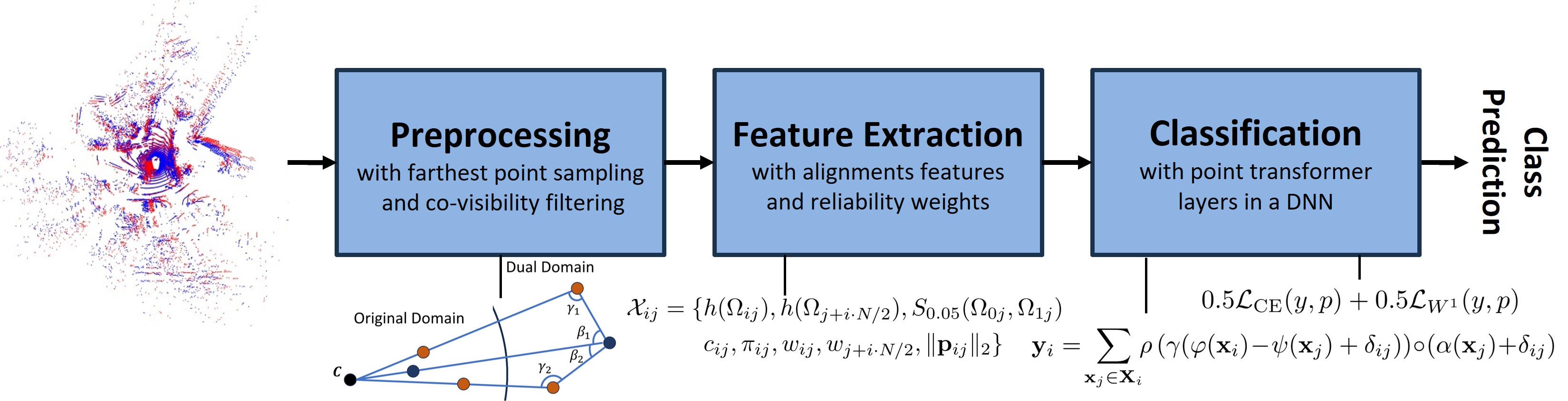}

    \caption{
    \textbf{Method overview}: A pair of registered point clouds are passed through preprocessing, feature extraction, and then a classifier that predicts the misalignment class. Key aspects of the corresponding modules are provided below the boxes. The image below the preprocessing box shows three points and their spherically flipped counterparts.
    }
    \label{fig:network_pipeline}
\end{figure}

FACT can be applied after registration to detect errors and validate the result. 
We generalize previous work by extending binary classification to multinomial classification to distinguish between different degrees of misalignment. 
This enhancement allows us to compare relative pose estimates at a finer level and assign confidence to the classification output.
With a more fine-grained estimate of the alignment quality, we can also prioritize which registrations to repair depending on the available resources. 
For example, if an agent is mapping an environment with lidar registrations, \ours~can provide detailed feedback on where the mapping is accurate and where improvements are needed.
Furthermore, when using multiple classes, small misclassifications lead to less drastic changes in system response, as neighboring classes represent similar levels of alignment quality.
For an overview of our method, see \cref{fig:network_pipeline}.

\section{Related Work}
\label{sec:related_work}

A classic way to reduce PCR errors is using {\it loop closure}, where the map is corrected whenever the agent revisits already mapped regions~\cite{ho2007detecting}. 
Droeschel \etal~\cite{droeschel2014local} instead suggest using mean map entropy to validate the transformation estimate between two point clouds, thus paving the way for automatic point cloud alignment validation. Entropy is proposed in~\cite{droeschel2014local} due to its ability to measure the \textit{sharpness} or \textit{crispness} of a map. 
However, entropy is also highly dependent on the scene structure, and as shown in~\cite{adolfsson2021coral}, a more distinct measure of alignment, called CorAl \cite{adolfsson2021coral}, is the {\it change in entropy} from the joint point cloud to the sum of the separate point clouds. 

\textit{Quality of point cloud registration}: The approaches \cite{almqvist2018learning,bogoslavskyi2017analyzing,nobili2018predicting} use one or a few features of a point cloud pair, such as spatial overlap or projection of 3D points to virtual image planes, to quickly estimate alignment quality. Other methods input their features into a logistic regression model to classify a point cloud pair as aligned or misaligned~\cite{yin2019failure,adolfsson2021coral,fujii2015detection}. Moreover, the works of~\cite{adolfsson2021coral,droeschel2014local,droeschel2018efficient} assess point cloud alignment quality using differential entropy. Conversely, \cite{akai2022detection} examines residual errors between points and the corresponding map in the context of Markov random fields with fully connected latent variables to detect localization failures.
Differently from others, the work of~\cite{camous2022deep} compares two deep learning architectures to solve the binary point cloud alignment problem.

\textit{Quality of point cloud maps}: For general point clouds, not limited to registered ones, Empir3D proposes four metrics to describe point cloud map quality~\cite{turkar2024empir3dframeworkmultidimensional}. They compare their suggested metrics with the Chamfer distance~\cite{wu2021densityaware}, Earth Mover's distance~\cite{wu2021densityaware}, and Hausdorff distance~\cite{Hausdorffdistance1993} and show superior performance. However, their suggested method requires a reference map typically being the ground truth, limiting its applicability for point cloud pairs.

\subsection{The CorAl Baseline}\label{sec:coral}
Now, we describe the baseline CorAl~\cite{adolfsson2021coral} and its two features, which we incorporate into FACT. For a point cloud pair ($\mathbf{P}_0, \mathbf{P}_1$), binary misalignment classification is based on \textit{differential entropy} measurements $h(\Omega_{ij})$ in local neighborhoods $\Omega_{ij}$ assuming a three-dimensional Gaussian distribution. Here, $i=0,1$ denotes which point cloud the point originates from and $j$ is the neighborhood index in each point cloud.
For each pair, the following two features are computed
\begin{equation}\label{eq:diff_entropy_feats}
        H_\mathrm{joint}(\mathbf P_{0,1}) = \frac{H(\mathbf P_{0,1})}{|\mathbf P_{0,1}|}, \quad
        H_\mathrm{sep}(\mathbf P_0, \mathbf P_1) = \frac{H(\mathbf P_0) + H(\mathbf P_1)}{|\mathbf P_{0,1}|}, 
\end{equation}
where $|\cdot|$ denotes cardinality and $\mathbf P_{0,1} = \mathbf{P}_0\cup\mathbf{P}_1$ is the joint point cloud. The sum of differential entropies is 
$H(\mathbf P_i) = \sum_{\Omega_{ij}\subseteq\mathbf P_i} h(\Omega_{ij})$
where 
\begin{equation}\label{eq:diff_eq_coral}
     h(\Omega_{ij}) = 0.5\ln\left((2\pi e)^3\det\Sigma(\Omega_{ij})\right), \Omega_{ij} = \left\{\mathbf p_{ik}\in\mathbf P_i : \Vert \mathbf p_{ij}-\mathbf p_{ik} \Vert_2 < r_{ij}\right\}
\end{equation}
Here, $\Sigma(\Omega_{ij})$ is the sample covariance of the point set $\Omega_{ij}$,
where $i\in\{0, 1, (0,1)\}$. The notation of joint local neighborhoods $\Omega_{(0,1),j}$ (\ie, considering points from both points clouds) is replaced with $\Omega_{j+i\cdot N/2}$ where $N/2$ is the number of neighborhoods per point cloud.
Additionally, \cite{adolfsson2021coral} introduces modifications to handle problematic scenarios, detailed in the paper. One important thing to mention is that the radius is chosen dynamically as $r_{ij} = d_{ij}\sin(\alpha)$ clipped to $[r_\mathrm{min}, r_\mathrm{max}]$, where $\alpha$ is the sensor's vertical angular resolution, and $d_{ij}=\lVert \mathbf{p}_{ij}\rVert_2$ is the distance from $p_{ij}$ to its corresponding lidar. 
Ultimately, the two features (see \cref{eq:diff_entropy_feats}) are fed to a logistic regression model for binary classification~\cite{james2013introduction}.

For training the classifier, CorAl systematically creates aligned and misaligned point cloud pairs in equal proportion. Aligned pairs are easily created using the ground truth transformations. Misaligned point cloud pairs are created by perturbing one of the point clouds in the pair, by adding an angular offset by rotating $\theta$ radians around the vertical axis of the sensor and a translational offset in the $(x,y)$ coordinates with the distance $e_d$ meters from the ground truth.

\section{Problem Formulation}

We define PCMC as the task of assigning a registration result into one out of $m$ bins, corresponding to different degrees of misalignment.
With multinomial, instead of binary classification, the registration estimates 
can be more precisely monitored, while simultaneously allowing the network to report uncertainty, in the form of several active outputs, which can be useful in downstream tasks.

\subsection{Data Generation}\label{sec:data_generation}
We use two data generation methods, and consequently, two categories of class definitions based on these methods. The first method is a straightforward generalization of the binary case used in CorAl~\cite{adolfsson2021coral} which allows us to compare performance with this baseline. The second method uses examples of actual point cloud registrations, partitioned using direct thresholding of their average {\it point transformation errors}. As the data generation methods and class definitions are different in the two cases (everything else is the same), we will train two separate networks.
The data used for training, validation, and testing are based on the nuScenes~\cite{nuscenes} and KITTI~\cite{geiger12} automotive datasets.

\subsubsection{Dataset with Generated Registration Errors:}\label{sec:fixed_error_classes}
Following CorAl~\cite{adolfsson2021coral}, we first generate a synthetic dataset comprising ten classes, each defined by the following rotation–translation pairs:
$(\theta, e_d)\in\{(0,0), (0.01, 0.1), \dots, (0.09, 0.9)\}$ (see \cref{sec:coral}).
Performing binary PCMC for large misalignment errors is significantly easier than for smaller ones~\cite{almqvist2018learning}. Thus, we focus on small errors (no larger than $5.2^{\circ}$ and $0.9$ m as seen in \cref{fig:intro_example}) that combined can propagate to large map-breaking errors through registration drift. It is important to note that a classifier that can separate these classes is not necessarily able to separate real registered lidar-scans, as these likely contain other types of errors.

\subsubsection{Dataset with PCR Errors:}\label{sec:pcr_error_classes}
Real point cloud registered data are created with registrations from the classical method ICP \textit{point-to-plane} by Open3D~\cite{Zhou2018} and the competitive method GeoTransformer~\cite{qin2022geometric}. On nuScenes, we use ICP, while on KITTI, we use GeoTransformer since the model is trained on this dataset. The datasets based on PCR errors have a more challenging data distribution to classify than the synthetically generated registration errors as it has continuous support.
However, this type of data corresponds to the intended use case, where we want to classify actual registrations.
For this data generation method, we create classes based on the average {\it point transformation error} $\epsilon$. This error uses the transformed point set ${\bf P}_1$ to define the metric, and measures the error for an estimated point
transformation $\tilde{\mathbf T}_{1\to 0}$ with respect to the ground truth 
transformation $\mathbf T_{1\to 0}$, as
\begin{equation}\label{eq:pt_trans_error}
    \epsilon = \frac{1}{|\mathbf{P}_1|}\sum_{k=1}^{|\mathbf{P}_1|} \Vert \mathbf T_{1\to 0}(\mathbf p_{1k}) - \tilde{ \mathbf T}_{1\to 0}(\mathbf p_{1k})\Vert_2.
\end{equation}
We define $5$ classes where the label $Y=k$ if $\epsilon\in I_k$. Here, $I_0 = [0, 0.03)$, $I_1 = [0.03, 0.10)$, $I_2 = [0.10, 0.25)$, $I_3 = [0.25, 0.50)$, and $I_4 = [0.50, \infty)$ with meter as the unit.

\section{Method}

Now, we present our method FACT (\textit{Feature-Aware Classification Transformer}).
FACT feeds features from optimal transport and information theory to a point transformer-based neural network that does multinomial classification.

\subsection{Preprocessing}\label{sec:preprocessing}
The first step is to preprocess the point cloud pair. During this step, we mainly perform co-visibility filtering and farthest point sampling.

\subsubsection{Co-visibility Filtering:}\label{sec:covisibility}
 PCMC should only be conducted in co-visible regions of the joint point cloud as in other regions alignment quality cannot be assessed.
 For each point, we divide the estimation of co-visibility into two tasks: a binary prediction (co-visible or hidden) and a soft prediction (degree of co-visibility).
 To solve the binary task, we use the {\it hidden point removal} operator~\cite{katz2007direct} to determine which points of a point cloud are visible from a given viewpoint (sensor position).
 For the points that are considered co-visible by the operator, we calculate a \textit{co-visibility score} building upon the framework of~\cite{katz2015visibility}. We explain our method for calculating the co-visibility scores in \cref{sec:app_covisibility_scores} and visualize how the scores can be a useful feature for detecting misalignments.

\subsubsection{Farthest Point Sampling and Neighborhoods:}
For a subset of the points in the joint point cloud, we calculate local features. To select a well-spread subset, we perform {\it farthest point sampling} (FPS)~\cite{eldar1997farthest} where $1024$ points are sampled from each point cloud.
Note that we use FPS to select which points to calculate features for, but when calculating the features, full point clouds are used.
Differently from CorAl~\cite{adolfsson2021coral}, we consider the distances to both sensor positions in the point cloud pair using the estimated transformation. This is motivated by that the density in the joint point cloud is dependent on the distances to both sensors.
We replace $d_{ij}\sin(\alpha)$ with $\sqrt{2}\sin(k\alpha)d_{ij}\tilde{d}_{ij}/(d_{ij}^2+\tilde{d}_{ij}^2)^{\frac{1}{2}}$ with $k=5$ to increase the neighborhood size faster than CorAl as we have sparser point clouds~\cite{adolfsson2021coral}. This together with
$r\in[0.5, 7.5]$ defines our neighborhoods' radii.

\subsection{Feature Extraction}\label{sec:feature_extraction}
Our classification network takes features $\mathcal X\in\mathbb R^{N\times D}$ and coordinates $\mathbf P \in\mathbb R^{N\times 3}$ as input, where $N=2048$ is the number of feature vectors and $D=8$ the feature dimensionality. 
We now present what each of these dimensions represent.
In the remainder of this section, we will consider a point $\mathbf{p}_{ij}$ selected via FPS, where $\mathbf{p}_{ij}\subseteq\Omega_{ij}\subseteq\Omega_{j+i\cdot N/2}\subseteq\mathbf{P}_{0,1}$. Like before, $\Omega_{ij}$ and $\Omega_{j+i\cdot N/2}$ are the point's neighborhoods in its own point cloud $\mathbf{P}_{i}$ and the joint point cloud $\mathbf{P}_{0,1}$, respectively.
As mentioned, we use differential entropy in our feature map. However, we use it locally instead of aggregating over the pair. \Cref{eq:diff_eq_coral} is used to define separate and joint local differential entropy measurements $h(\Omega_{ij})$ and $h(\Omega_{j+i\cdot N/2})$, which are our two first feature dimensions.

\subsubsection{Sinkhorn Divergence:}\label{sec:sinkhorn}
Sinkhorn divergence is an extension of Sinkhorn distance with improved convergence and bias properties~\cite{genevay2018learning,feydy2019interpolating}.
Sinkhorn distance is an entropic regularization approach to the optimal transport problem~\cite{cuturi2013sinkhorn}. 
\textit{Entropic regularized optimal transport} is defined as
\begin{equation}\label{eq:sinkhorn_dist}
    W_\varepsilon(\alpha, \beta) = \min_{\pi\in\Pi(\alpha, \beta)}\int_{\mathcal{X}\times \mathcal{Y}}\left(c(x,y) + \varepsilon\ln\left(\frac{\mathrm{d}\pi(x,y)}{\mathrm{d}\alpha(x) \mathrm{d}\beta(y)}\right)\right)\mathrm{d}\pi(x,y),
\end{equation}
where $\alpha(x), \beta(y)$ are probability measures, $c(x,y)$ is the cost of moving a unit of mass from source $x$ to target $y$, $\pi(x,y)$ is the amount of mass moved, $\Pi(\alpha, \beta)$ denotes the set of transport plans ensuring all mass is transferred from source to target without negativity, and $\varepsilon$ is the regularization parameter~\cite{genevay2019sample,feydy2019interpolating}. \textit{Sinkhorn divergence}, proposed by~\cite{genevay2018learning,salimans2018improving}, is our third feature and it can be written as
\begin{equation}\label{eq:sinkhorn_div}
    S_\varepsilon(\alpha,\beta) = W_\varepsilon(\alpha,\beta) - 0.5W_\varepsilon(\alpha,\alpha)-0.5W_\varepsilon(\beta,\beta). 
\end{equation}

\subsubsection{Reliability Weights:}
Here, we describe our five reliability weights.
The first reliability weight is the \textit{co-visibility scores} proposed in \cref{sec:covisibility}.
The second one is a binary feature, $\pi_{ij}$, encoding which point cloud a point originates from.
Two other reliability weights we use are $w_{ij} = |\Omega_{ij}|/|\mathbf{P}_i|$ and $w_{j+i\cdot N/2} = |\Omega_{j+i\cdot N/2}|/|\mathbf{P}_{0,1}|$. They are used to inform the classifier of the proportion of the separate and joint point clouds in the local feature extraction. If few points are used, the measurements will be noisy and thus less reliable. On the contrary, there will be more measurements from denser regions of the point cloud; however, the farthest point sampling reduces that sampling imbalance.
Lastly, to enable the network to weigh parts of 3D space differently without overfitting the training data, we add $\Vert \mathbf{p}_{ij}\Vert_2$ as a feature.

\subsection{Classification}\label{sec:classification}
In this work, we leverage the advancements of the transformer architecture adapted to point clouds as done in~\mbox{\cite{guo2021pct,zhao2021point}}. The self-attention operator should be particularly useful for point cloud data as the operator is invariant to permutations and the cardinality of the input~\cite{zhao2021point}.
We extend the Point Transformer classification framework~\cite{zhao2021point} to perform PCMC with our feature map $\mathcal{X}$ as input. Here, we explain our loss function and refer to \cref{sec:app_classification_architecture} for details about the point transformer layer and the network architecture.

\subsubsection{Loss Function:}\label{sec:loss}
Point Transformer~\cite{zhao2021point} does shape classification on ModelNet40~\cite{wu20153d} with a cross-entropy loss, see \eg, the reimplementation~\cite{pointtransformers}. This is reasonable because there is no inherent order between the classes, \ie, mispredictions should be penalized equally. However, for our task, it is more suitable to do \textit{regression-by-classification}~\cite{torgo1996regression}, such that the loss considers the inherent order of the classes, and treats near misses as less severe. To do this, we use a \textit{Wasserstein loss}, which has a simple formulation in one dimension~\cite{thorarinsdottir13}. The discrete Wasserstein loss between the ground truth one-hot vector $y$ and the predicted distribution $p$ in one dimension is given by
\begin{equation}
    \mathcal{L}_{W^1}(y,p) = \frac{1}{n}\sum_{i=0}^{n-1}\sum_{k=0}^{m-1}|Y_{ik} - P_{ik}| = \frac{1}{n}\sum_{i=0}^{n-1}\sum_{k=0}^{m-1}|\sum_{l=0}^ky_{il} - \sum_{l=0}^kp_{il}|,
\end{equation}
where $Y_{ik}$, $P_{ik}$ are CDFs. However, the $W_1$ loss combined with softmax can lead to vanishing gradients impairing learning updates~\cite{xiong2024hinge}. Thus, we suggest combining the $W_1$ with the cross-entropy loss, $\mathcal{L}_\mathrm{CE}(y, p)$, to get
\begin{equation}\label{eq:loss}
        \mathcal{L}(y,p) = 0.5\mathcal{L}_\mathrm{CE}(y,p) + 0.5 \mathcal{L}_{W^1}(y,p).
\end{equation}

\section{Experiments}
To test FACT, we compare with the contender CorAl~\cite{adolfsson2021coral} on synthetically perturbed data.
Then, we test FACT on ICP \textit{point-to-plane} registered data from nuScenes and then GeoTransformer registered data from KITTI.
After that, we show that registration residuals and standard point cloud quality metrics fail to accurately quantify alignment errors.
We conclude with a demonstration of how FACT can be used for point cloud map correction by highlighting which point clouds are incorrectly registered so that we in this case can manually correct them. For more training details see \cref{sec:training_details}.

\subsection{Classification of Generated Errors}\label{sec:multinomial_classification}
Here, we show that FACT generalizes prior work by handling more than two cases.
\ours~is trained on ten classes but tested on two of them. This not only makes \ours~more general but also allows us to employ a simple mapping rule.
The mapping rule can be defined differently based on which misclassifications are most problematic, false negatives or false positives.
The class-agnostic approach is to map predictions that do not belong to any of the two binary classes to the closest one. If the distances to the two classes are equal, the prediction can be distributed equally between the binary classes. Using this idea, we obtain \cref{tab:binary_comparison} showing that \ours, with this mapping rule employed, achieves great performance, and surpasses the accuracy of CorAl by a large margin. Additionally, using a multinomial classifier for binary classification enables incorporating confidence in the predictions. For example, prediction uncertainty could be based on the entropy of the network output vector.
Moreover, we believe that it is hard to generalize from pre-defined perturbations to real data and thus focus all other experiments on real point cloud registrations.

\begin{table}[ht]
    \centering
    \caption{Classification accuracy for CorAl and FACT on two classification tasks. %$(\theta, e_d)=(0.01, 0.1)$ and $(\theta, e_d)=(0.03, 0.3)$.
    }
    \begin{tabular}{@{}>{\centering\arraybackslash}p{2.7cm}>{\centering\arraybackslash}p{2.7cm}>{\centering\arraybackslash}p{2.7cm}@{}}
        \toprule
        \textbf{Task} & \textbf{CorAl~\cite{adolfsson2021coral}} & \textbf{Mapped FACT} \\ 
        \midrule
        $(\theta, e_d)=(0.01, 0.1)$ & 75.3\% & \textbf{97.4\%} \\
        $(\theta, e_d)=(0.03, 0.3)$ & 95.6\% & \textbf{100.0\%} \\
        \bottomrule
    \end{tabular}
    \label{tab:binary_comparison}
\end{table}

\subsection{Multinomial Classification of PCR Errors}
Now, we demonstrate how FACT performs when classifying actual registered point cloud pairs. The setup here is always a point cloud pair and a relative transformation estimate by either ICP-p2l or GeoTransformer giving an average \textit{point transformation error} that we try to predict.

\begin{figure}
    \centering
    \begin{minipage}[b]{0.5\columnwidth}
    \centering
    \includegraphics[width=41.5mm, trim={0 0 0 0.2cm}, clip]{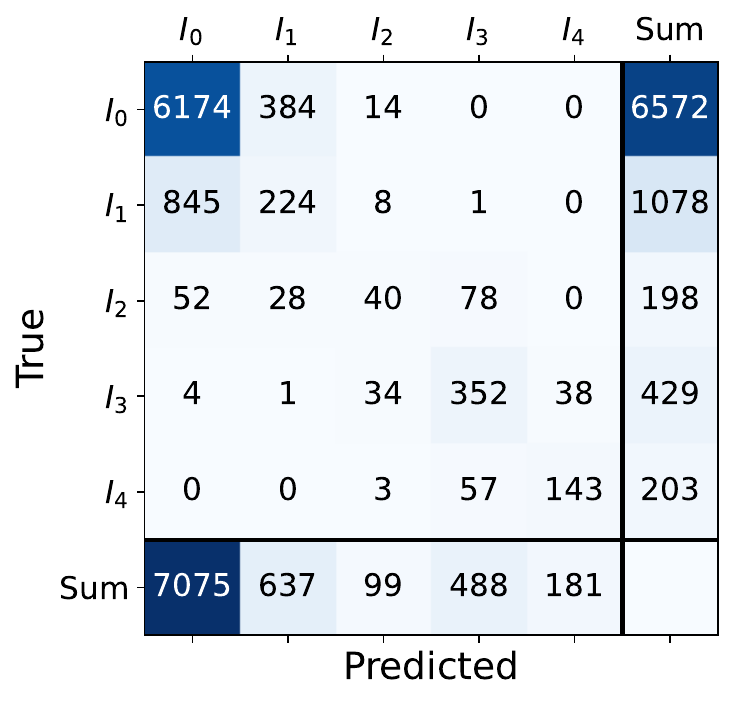}
    \subcaption{Regression-by-classification}
    \label{fig:conf_matrix_34000_regression-by-classification}
    \end{minipage}%
    \begin{minipage}[b]{0.5\columnwidth}
    \centering
    \includegraphics[width=41.5mm, trim={0 0 0 0.2cm}, clip]{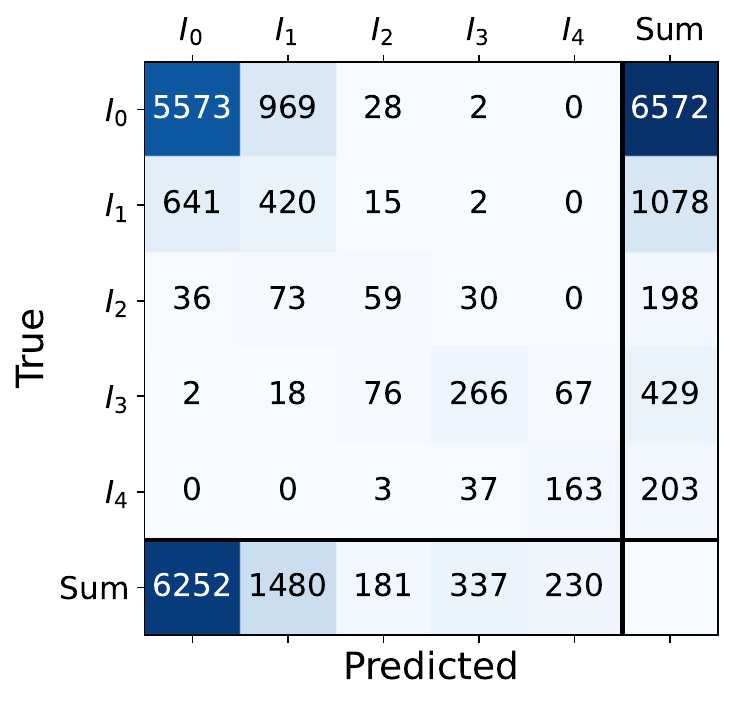}
    \subcaption{Regression}
    \label{fig:conf_matrix_34000_regression}
    \end{minipage}%
    \caption{
    %Confusion matrices for the ICP-p2l registration-based test dataset on nuScenes created with the procedure described in \cref{sec:pcr_error_classes}.
    Confusion matrices for the ICP-p2l registration-based test dataset on nuScenes.
    The left figure is for our proposed loss described in \cref{sec:loss} and the right figure is when our loss is replaced by an ordinary mean squared error loss.}
    \label{fig:conf_matrix_34000}
\end{figure}

\begin{table}[ht]
    \centering
    \caption{Comparison between regression-by-classification (RbC) and regression. $\xi_k$ represents the fraction of samples where the predicted label is at least $k$ classes away from the true label.}
    \begin{tabular}{@{}lcccc@{}}
        \toprule
        & $\xi_1$ & $\xi_2$ & $\xi_3$ & $\xi_4$ \\ 
        \midrule
        \textbf{RbC}         & \textbf{18.24\%} & \textbf{0.88\%} & \textbf{0.05\%} & \textbf{0.00\%} \\
        \textbf{Regression}  & 23.57\%          & 1.07\%          & \textbf{0.05\%} & \textbf{0.00\%} \\
        \bottomrule
    \end{tabular}
    \label{tab:comparison_rbc_vs_reg}
\end{table}

\subsubsection{Classifying ICP Errors in nuScenes:}\label{sec:icp_nuscenes_exp}
Our predictions and the ground truth labels for the registration-based test dataset of 8,480 point cloud pairs can be seen in \cref{fig:conf_matrix_34000_regression-by-classification}. To show the benefits of the regression-by-classification methodology we use, the results obtained when replacing the regression-by-classification head and loss with a regression head and loss are also presented. In the regression variant, we project to one dimension instead of $N_\textrm{classes}$ dimensions in the final MLP, and then apply the softplus activation, $\ln(1+e^x)$, to ensure positivity and lastly insert the value into a mean squared error loss.
Both alternatives can effectively separate aligned from misaligned point cloud pairs. To make the difference between the alternatives clearer, we compare them in \cref{tab:comparison_rbc_vs_reg}. We see that regression-by-classification is on-par with, or better than direct regression for all four error categories, which is the property we value the most. Thus, we conclude that regression-by-classification outperforms regression, but that both variants give great results as severe errors made by the classifiers are very rare. FACT can thereby also classify real point cloud registration errors successfully.

\subsubsection{Classifying GeoTransformer Errors in KITTI:}\label{sec:supp_geo_kitti}

\begin{table}[ht]
    \centering
    \caption{Confusion matrix for the GeoTransformer registration-based test dataset on KITTI. Classes 0 and 1 have been collapsed into one class.}
    \begin{tabular}{l >{\centering\arraybackslash}p{0.80cm} >{\centering\arraybackslash}p{0.80cm} >{\centering\arraybackslash}p{0.80cm} >{\centering\arraybackslash}p{0.80cm}}
        \toprule
        \textbf{Predicted Class}& $I_0\cup I_1$ & $I_2$ & $I_3$ & $I_4$ \\ 
        \midrule
        \textbf{True Class }$I_0\cup I_1$ & 392 & 11 & 0 & 0 \\
        \textbf{True Class }$I_2$ & 59  & 19 & 3 & 0 \\
        \textbf{True Class }$I_3$ & 0   & 3  & 5 & 1 \\
        \textbf{True Class }$I_4$ & 0   & 1  & 4 & 2 \\
        \bottomrule
    \end{tabular}
    \label{tab:conf_matrix_geotrans}
\end{table}

Here, we study point cloud pairs registered with GeoTransformer which is a state-of-the-art registration method~\cite{qin2022geometric}. We split the 11 scenes into training, validation, and test sets as in~\cite{qin2022geometric}, with a total of roughly 2,000 point cloud pairs. To obtain better performance, we used our network trained on nuScenes with ICP to initialize the network weights and then fine-tuned on the GeoTransformer registered KITTI pairs. We noticed that GeoTransformer had a more difficult time estimating relative poses with an average point transformation error $\epsilon<0.03$ compared to ICP. Hence, we collapsed classes 0 and 1 from before to one class after training. 
 
 \Cref{tab:conf_matrix_geotrans} shows how FACT manages to classify GeoTransformer errors. We see that the classifier only once makes an error that does not belong to a neighboring class of the ground truth class. With this information, one could for example re-register the point cloud predicted as class 2 and 3 and likely be able to remove most of the errors that a point cloud map contains. Even if the results are good here, we do see a degradation in performance compared to ICP on nuScenes, likely due to the lack of training data in the KITTI dataset.

\subsection{Other Methods for Misalignment Detection}\label{sec:app_other_methods}
We now show how alternatives for misalignment detection like ICP registration residuals and the Hausdorff distance fail to predict the registration error.

\subsubsection{Misalignment Detection with ICP Residuals:}\label{sec:app_reg_residuals}
A possible way to correct misalignments is by using registration residuals. Nonetheless, we have not found this satisfactory. Most deep learning methods do not have a residual metric for detecting possible errors.
However, ICP \textit{point-to-plane} from Open3D~\cite{Zhou2018} used in many of the experiments returns a fitness and an inlier root mean squared error (RMSE) score. The fitness measures the overlapping area of the inlier correspondences and the points in the target point cloud (higher is better). The inlier RMSE measures the RMSE of the inlier correspondences (lower is better). We investigated the applicability of both these metrics for predicting the misalignment class. Our experiments on 1,060 point cloud pairs showed that the correlation between the ground truth label (0 to 4) and the fitness was -0.15 whereas it was 0.065 between the ground truth label and the inlier RMSE. On the contrary, FACT's predictions had a correlation of 0.87 with the ground truth labels. In conclusion, the ICP registration residuals offered no effective use for PCMC, serving only as very weak indicators of misalignment.

\subsubsection{Misalignment Detection with Standard Metrics:}\label{sec:misalignment_detection_with_common_metrics}
\begin{figure*}
    \centering
    \includegraphics[width=1.0\linewidth, trim={0 0 0 0.0cm}, clip]{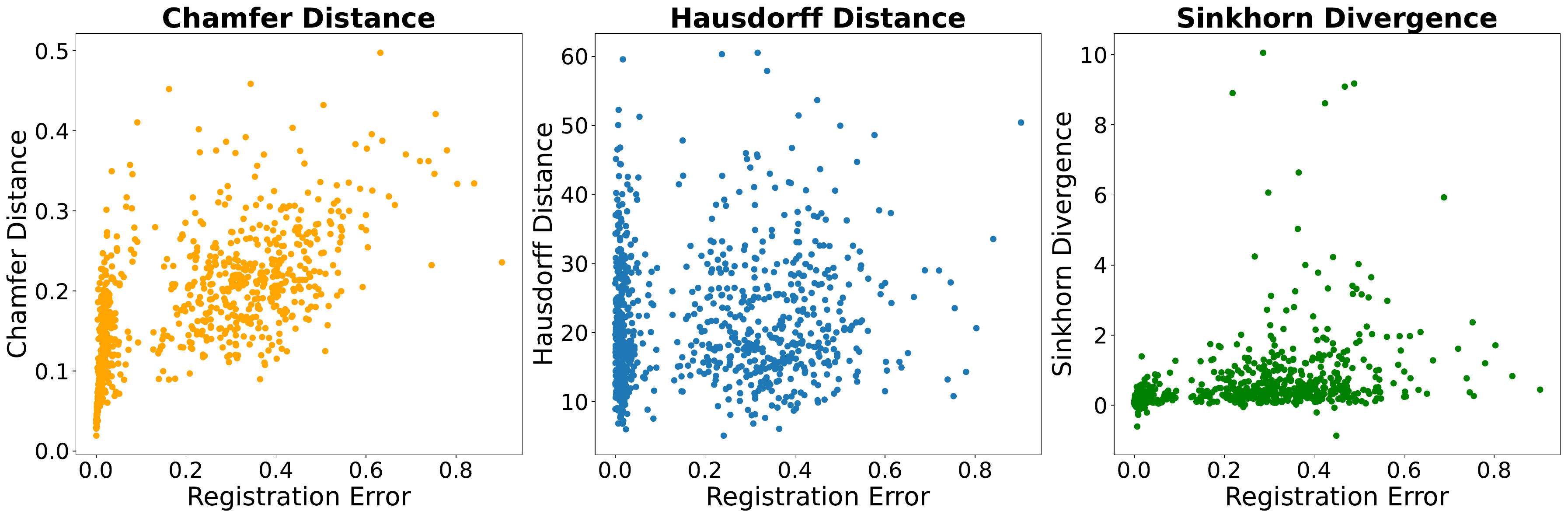}
    \caption{
    %Chamfer distance, Hausdorff distance, and Sinkhorn divergence versus the registration error (average {\it point transformation error} in \cref{eq:pt_trans_error}) for 850 point cloud pairs from 850 different scenes in nuScenes registered with ICP-p2l.
    Chamfer distance, Hausdorff distance, and Sinkhorn divergence vs. the registration error ($\epsilon$) for 850 nuScenes point cloud pairs registered with ICP-p2l.}
    \label{fig:metrics_vs_registration_error}
\end{figure*}
Methods like Chamfer Distance, Hausdorff Distance, and Earth Mover's Distance (EMD) are often used as metrics for point cloud alignment quality~\cite{turkar2024empir3dframeworkmultidimensional}. However, we find that none of these metrics are reliable enough by themselves for estimating if points clouds are correctly aligned. When studying this, we registered 850 point cloud pairs taken from different scenes in nuScenes~\cite{nuscenes} using ICP. We computed the Chamfer distance~\cite{wu2021densityaware}, the Hausdorff distance~\cite{Hausdorffdistance1993}, and the Sinkhorn divergence~\cite{feydy2019interpolating} between the point sets after registration.
As seen in \cref{fig:metrics_vs_registration_error}, it is hard to draw a precise conclusion of the registration error by looking at one metric alone.
However, if the Chamfer distance is below 0.05, the registration error is typically very small; whereas if the Sinkhorn divergence is above 3, the registration error is not small.
The metrics will only help in deducing if a small subset of the pairs are aligned implying that a better method for distinguishing between aligned and misaligned pairs is needed.
Note that the metrics here are calculated over the full point clouds while FACT aggregates metrics calculated in local neighborhoods.

\begin{figure*}[!htbp]
    \centering
    \begin{minipage}[b]{0.25\textwidth}
        \centering
        \includegraphics[width=1\linewidth, trim={0 0 0 0.0cm}, clip]{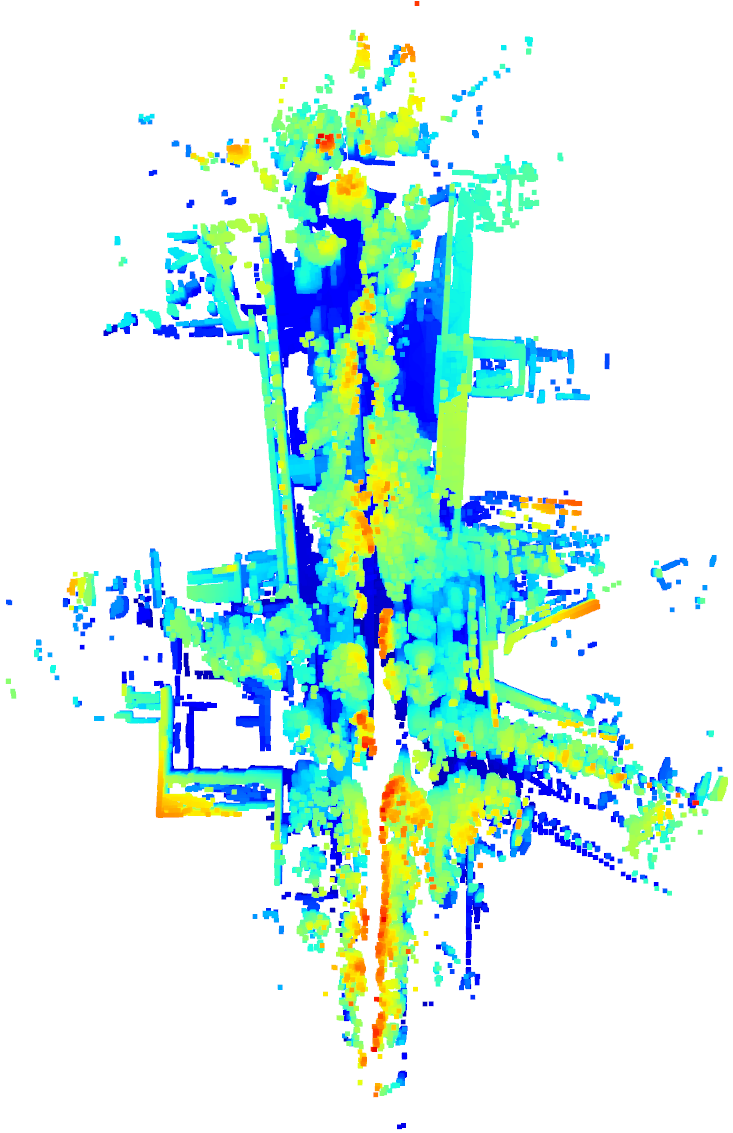}
        \subcaption{Registered scene}\label{fig:gt_corrected_a}
    \end{minipage}%
    \begin{minipage}[b]{0.25\textwidth}
        \centering
        \includegraphics[width=1\linewidth, trim={0 0 0 0.0cm}, clip]{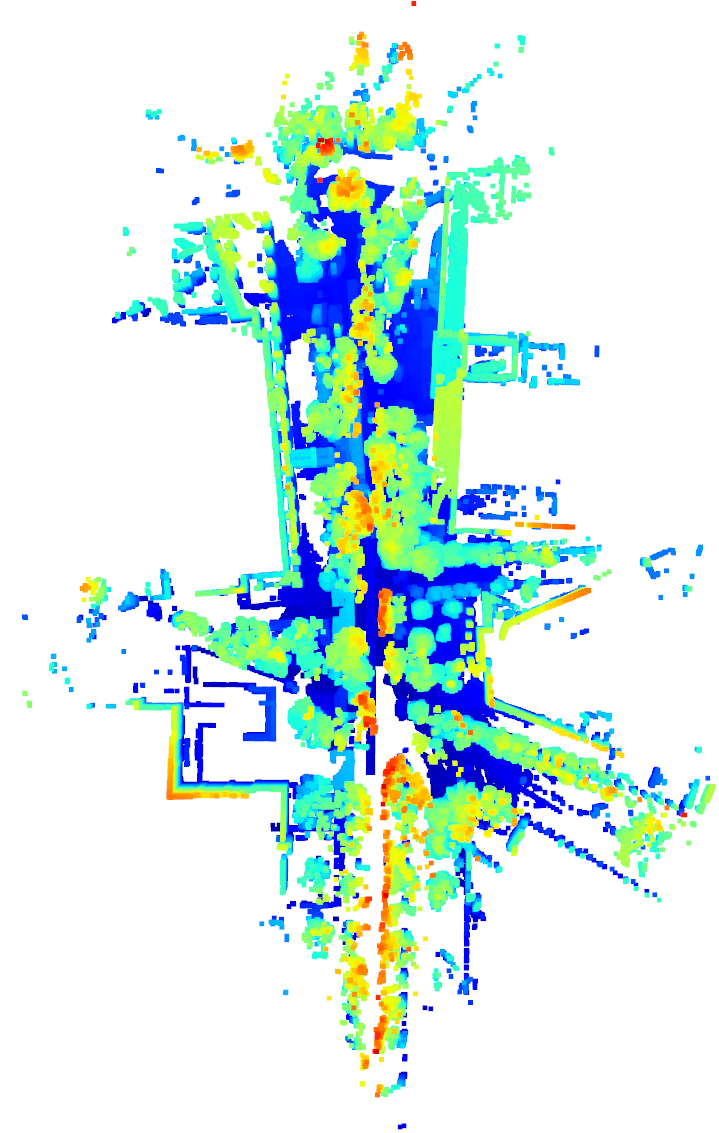}
        \subcaption{Repaired scene}
        \label{fig:gt_corrected_b}
    \end{minipage}%
    \begin{minipage}[b]{0.25\textwidth}
        \centering
        \includegraphics[width=1\linewidth, trim={0 0 0 0.0cm}, clip]{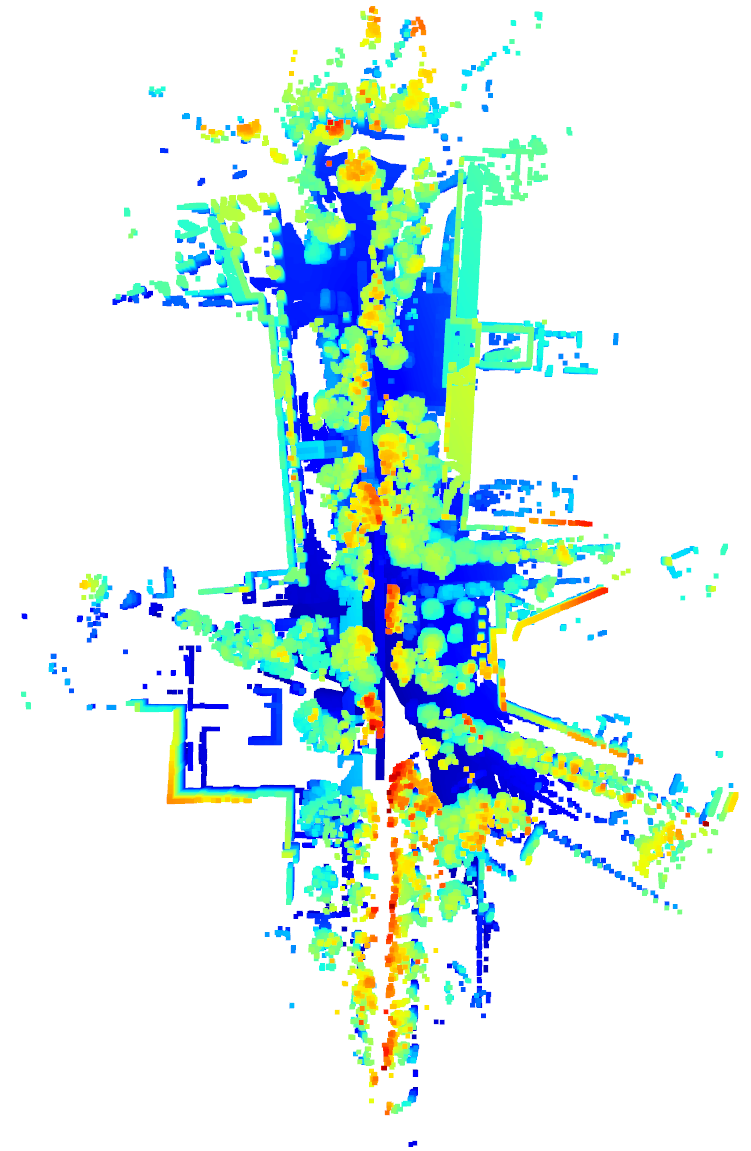}
        \subcaption{Ground truth}
        \label{fig:gt_corrected_c}
    \end{minipage}%
    \begin{minipage}[b]{0.25\textwidth}
        \centering
        \raisebox{0.75cm}{\includegraphics[width=1\linewidth, trim={0 0 0 0.0cm}, clip]{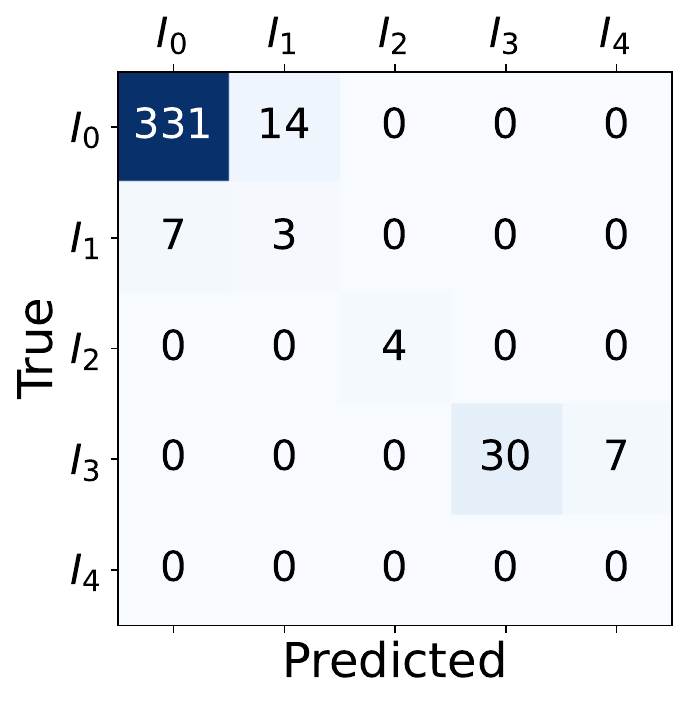}}
        \subcaption{Confusion matrix}
        \label{fig:gt_corrected_d}
    \end{minipage}%
\caption{
In (a), a test scene has been registered with ICP-p2l. FACT correctly detects the 37 relative poses (9.3\% of all) with an average {\it point transformation error} greater than 0.25 meters (see (d)). Setting these relative poses to the ground truth results in (b). The point map color represents height.}
\label{fig:gt_corrected}
\end{figure*}

\subsection{Point Cloud Map Correction}\label{sec:experiments_map_corr}
With accurate predictions of which point cloud pairs are misaligned, measures to correct point cloud maps can be taken.
For predicted misalignments, manual correction is commonly done in the offline mapping context (as part of the quality assurance process), whereas in online mapping, an agent can revisit misaligned areas to improve the map.
As an example, we show the equivalent of manual correction of detected misalignments by setting the samples predicted as class 3 or 4 to the ground truth relative pose. This is done on a full scene registered with ICP-p2l on nuScenes (see \cref{fig:gt_corrected}). By only correcting 9.3\% of the relative poses, the point cloud map goes from being largely wrong to nearly identical to the ground truth map, demonstrating the usefulness of FACT.

\section{Conclusion}
This paper demonstrates that the point cloud misalignment classification task can effectively be solved with a feature-aware point transformer-based neural network. Specifically, we show that FACT can classify synthetically perturbed point cloud pairs as well as both ICP and GeoTransformer registered point clouds. Our method outperforms CorAl by a large margin with respect to misalignment classification accuracy and generalizes previous methods by performing multinomial instead of binary alignment classification.
We hope FACT will find use as a means for self-supervision in learning settings, as a quality metric in model evaluation, and uncertainty representation in point cloud registration pipelines.

\begin{credits}
\subsubsection{\ackname} This work was supported by the strategic research environment ELLIIT funded by the Swedish government, the Wallenberg AI, Autonomous Systems and Software Program (WASP) funded by the Knut and Alice Wallenberg Foundation, by Vinnova, project \textnumero $2023-02694$, and Formas project \textnumero 2023-00082. The computations were %partially
enabled by the Berzelius resource provided by the Knut and Alice Wallenberg Foundation at the National Supercomputer Centre.
%\subsubsection{\ackname} Placeholder.

\subsubsection{\discintname}
The authors have no competing interests to declare that are
relevant to the content of this article.
\end{credits}
\clearpage
\bibliographystyle{splncs04}
\bibliography{main}

\clearpage
\setcounter{page}{1}
\maketitlesupplementary

\appendix

\section{Network Details}\label{sec:app_classification_architecture}

Zhao \etal~\cite{zhao2021point} define their point transformer layer as
\begin{equation}\label{eq:point_trans_layer}
    \mathbf y_i = \sum_{\mathbf x_j\in \mathbf X_i}\rho\left(\gamma(\varphi(\mathbf x_i)-\psi(\mathbf x_j)+\theta(\mathbf p_i-\mathbf p_j)\right))\circ(\alpha(\mathbf x_j)+\theta(\mathbf p_i-\mathbf p_j))
\end{equation}
where $\mathbf X_i \subseteq \mathbf X$ is the set of feature vectors corresponding to the $k$ nearest neighbors of $\mathbf p_i$, $\rho(\cdot)$ is a normalized softmax operator, and $\varphi(\cdot), \psi(\cdot), \alpha(\cdot)$ are the query, key, and value vectors. Moreover, $\theta(\mathbf p_i-\mathbf p_j)$ is a relative position encoding and $\theta(\cdot), \gamma(\cdot)$ are MLPs. Additionally, $\circ$ denotes the Hadamard product.

\begin{figure*}
    \centering
    \begin{minipage}{1.00\textwidth}
        \centering
        \includegraphics[width=1\linewidth, clip]{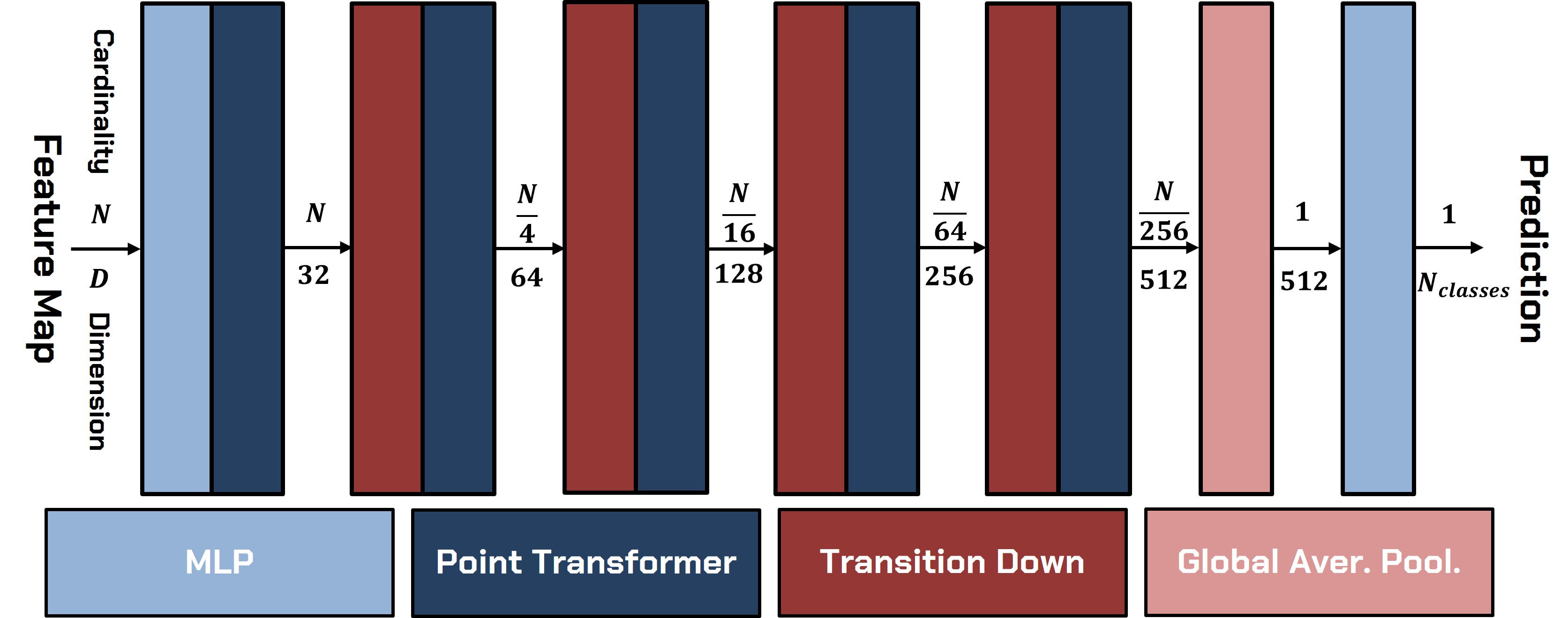}
    \end{minipage}%
    \caption{The classification architecture presented in~\cite{zhao2021point} (adapted from the corresponding figure in the paper). A $D$ dimensional point cloud with $N$ points is input into five double-blocks that encode features and progressively downsample the point cloud. Thereafter, the spatial dimension is pooled and fed through an MLP which outputs logits for the given classes.}
    \label{fig:point_trans_class_arch}
\end{figure*}

We use the architecture from Point Transformer~\cite{zhao2021point} (shown in \cref{fig:point_trans_class_arch}) and the code from~\cite{pointtransformers}. In brief, the \textit{point transformer} block consists of a linear layer, followed by a point transformer layer, and another linear layer, with a residual connection.
The \textit{transition down} block consists of {\it farthest point sampling} (FPS) with a rate of four, a $k$NN graph (based on the input point cloud to this block) fed into a {\it multi-layer perceptron} (MLP) with a linear layer, batch normalization, and ReLU followed by max pooling over the neighbors.
As a result, the block reduces the spatial dimension by four times. Finally, all the features are fed through a global average pooling layer over the remaining spatial dimensions followed by an MLP with three linear layers and two ReLU activations. This outputs class logits, where the max score is chosen as the class prediction. For more details, please refer to the original publication~\cite{zhao2021point}. 

\section{Visualization of Co-visibility Scores}\label{sec:app_covisibility_scores}
 When explaining our method, we use the image below the preprocessing box in \cref{fig:network_pipeline} as a reference. 
 If the blue point lies on the convex hull of the points in the dual domain, we investigate its degree of visibility. 
 The blue point is more likely to be visible if it lies in front of its neighbors (orange points) w.r.t.\ the viewpoint $\mathbf{c}$ in the original domain. 
 This corresponds to $\gamma_k - \beta_k$ being large. Different from 2D, each vertex (inverted point) on the convex hull in 3D will have many connected vertices. 
 Hence, we want the average, $\nu$, of $\{\gamma_k - \beta_k\}_{k=1}^M$ to be large, where $M$ is the number of neighbors.
 To calculate the \textit{co-visibility scores}, we first gather all vertices and the viewpoint (which technically also is a vertex). 
We then calculate every triangle side length (see \cref{fig:network_pipeline}) and use the law of cosines to get all ($\gamma$, $\beta$) pairs giving a $\nu$ for each vertex. 
Lastly, $\nu$ is min-max normalized over its point cloud to lie within $[0, 1]$ giving the co-visibility scores. 

\begin{figure*}
    \centering
    \begin{minipage}{0.46\textwidth}
        \centering
        \includegraphics[width=0.85\linewidth, trim={0 0 1.6cm 0.775cm}, clip]{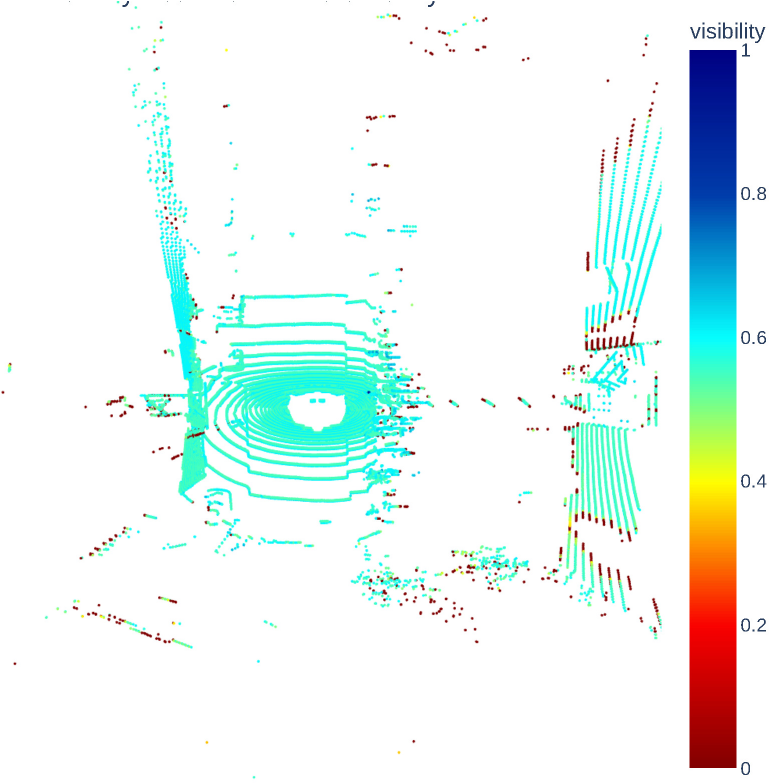}
    \end{minipage}%
    \begin{minipage}{0.54\textwidth}
        \centering
        \includegraphics[width=0.85\linewidth, trim={0 0 0 0.775cm}, clip]{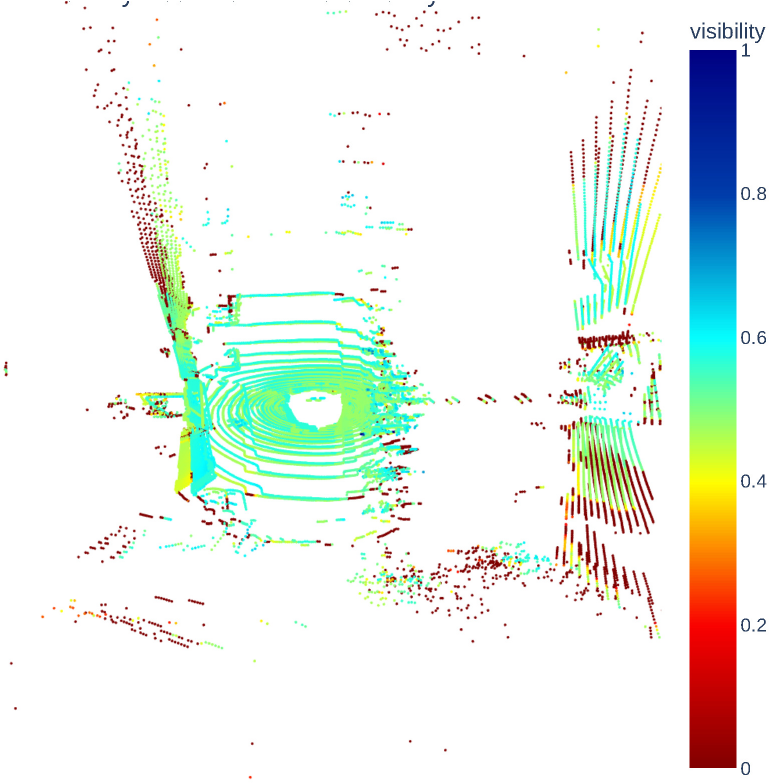}
    \end{minipage}%
    \vspace{-2.5mm}
   \caption{Co-visibility scores for two variants of one point cloud pair from nuScenes~\cite{nuscenes}. The left figure shows an aligned pair and in the right figure, one point cloud is perturbed with $(\theta, e_d)=(0.05, 0.5)$ (see \cref{sec:coral}).}
\label{fig:nsc_spherical_flipping}
\end{figure*}

In \cref{fig:nsc_spherical_flipping}, we demonstrate that the co-visibility score effectively describes the actual co-visibility of joint point clouds. For instance, the left part of the right figure correctly shows a significant co-visibility discrepancy between points. 

\section{Training Details}\label{sec:training_details}
We use ADAM~\cite{kingma2014adam} as optimizer with a weight decay of $10^{-4}$ and a step-based learning rate scheduler. For the Sinkhorn divergence, we use $c(x, y) = \frac{1}{2}\lVert x -  y \rVert^2_2$ as cost function in \textit{geomloss}~\cite{feydy2019interpolating}. We split the data into 60\% training data, 15\% validation data, and 25\% test data. On an NVIDIA GeForce RTX 4090, feature extraction takes approximately 1.5 seconds while classification inference takes roughly 5 milliseconds on nuScenes. In our longest training session, feature extraction took 14 hours and classification training took 13 hours.

\end{document}